\documentclass[runningheads]{llncs}
\usepackage[T1]{fontenc}
\usepackage{graphicx}
\usepackage{amsmath}
\usepackage{amssymb}
\usepackage{booktabs}
\usepackage{siunitx}
\usepackage{paralist}
\newcommand{\dset}[1]{\textsf{#1}}
\begin{document}
\title{ELADO: Elliptic PDE Assessment Datasets\\ for Operator Learning}
\author{Frank Ehebrecht\inst{1,2}\orcidID{0009-0001-5680-0256} \and
Toni Scharle\inst{2}\orcidID{0009-0007-8306-1042} \and
Martin Atzmueller\inst{1,3}\orcidID{0000-0002-2480-6901}}
\authorrunning{F. Ehebrecht et al.}
\institute{
Osnabr\"uck University, Semantic Information Systems Group, Osnabr\"uck, Germany \and
ROSEN Technology and Research Center GmbH, Lingen, Germany \and
German Research Center for Artificial Intelligence (DFKI), Osnabr\"uck, Germany }
\maketitle
\begin{abstract}
We introduce ELADO (Elliptic PDE Assessment Datasets for Operator Learning), a systematic benchmark suite constructed to show and quantify failure modes of neural operator architectures when learning solution operators of elliptic PDEs.
While the benchmarks of existing datasets focus on average case performance, the ELADO datasets are constructed to highlight challenges that arise naturally in elliptic PDE problems.

In particular, we construct several datasets built around Poisson's equation and the Helmholtz equation, each with non-constant coefficients. 
We define a controllable data-generating process to create datasets, that are designed to isolate a distinct source of difficulty. Specifically, these are
(1) heavy-tailed solution distributions arising from light-tailed coefficient field distributions,
(2) spectral distribution shift of the input data,  
(3) heavy-tailed distributions in the frequency domain of solutions, arising from light-tailed coefficient field distributions,
(4) input sensitivity of learned operators, quantified by an empirical local Lipschitz analysis, and
(5) the effect of input signal complexity on prediction accuracy under controlled amplitude normalization.

We evaluate several neural operator architectures across all datasets and show that heavy-tailed targets, spectral shift, and input sensitivity each cause substantial degradation of the prediction accuracy that standard datasets and metrics (e.g., the mean relative $L^2$ error) may obscure.
\end{abstract}
\section{Introduction}
Operator Learning has emerged as a new promising field for approximating PDE solution operators efficiently \cite{Kovachki2021NeuralOL,lu2019deeponet,li2020fourier,azizzadenesheli2024neural}. Despite considerable advances in the field, current publications rely heavily on a small set of standard datasets \cite{li2020fourier,li2023fourier,takamoto2024pdebenchextensivebenchmarkscientific,gupta2022towards}, commonly evaluated on aggregated measures (often variants of the relative $L^2$ error). Yet a uniform distribution over PDE inputs does not, in general, induce a uniform distribution over solution characteristics: heavy-tailed output norms, and heterogeneous spectral content can all concentrate difficulty in a tiny portion of the data that aggregate metrics effectively mask. These effects may be directly caused by the properties of a solution operator itself. Because neural operators learn a data-dependent surrogate rather than the abstract solution operator, the training distribution directly governs the quality of the learned map.
\setcounter{footnote}{0}
In this paper, we introduce ELADO\footnote{https://zenodo.org/records/20738464} (Elliptic PDE Assessment Datasets for Operator Learning), a diagnostic benchmark suite for elliptic PDE operator learning constructed via a controllable data-generating process in order to isolate distinct sources of difficulty -- heavy-tailed targets, spectral distribution shift, heavy-tailed frequency content, amplified input-output sensitivity, signal-complexity -- and we demonstrate across multiple neural operator architectures, that these effects can substantially degrade predictive accuracy in ways that standard datasets and single-number metrics may mask.
\section{Related Work}
Neural operator learning builds on the idea that maps between function spaces can be approximated by neural networks from discretized observations \cite{chen1995universal}. The pioneering DeepONet framework \cite{lu2019deeponet} introduced the branch--trunk decomposition, encoding input function samples and query coordinates in separate networks. The original formulation assumes a fixed set of input sensors shared across all samples, motivating a range of extensions that preserve the branch--trunk architecture while improving a variety of aspects \cite{liu2021multiscale,liu2022causality}.

In parallel, the Fourier Neural Operator (FNO) architecture \cite{li2020fourier} proposed a new kind of layer based on learning Fourier modes, demonstrating significantly improved accuracies and resolution-transfer behavior. Subsequent work extended this model class to more general geometries and irregular discretizations~\cite{li2023fourier,li2020multipole}.

On the benchmarking side, \cite{li2020fourier} and \cite{li2023fourier} set some of the de-facto standard datasets in the operator learning community. Broad suites such as PDEBench or PDEArena standardize across many PDEs \cite{takamoto2024pdebenchextensivebenchmarkscientific,gupta2022towards}, but they are not designed to isolate PDE-specific failure modes under controlled tail behavior, spectral-shift, and input sensitivity. ELADO complements these benchmarks by providing targeted elliptic PDE datasets that quantify such failure modes systematically.
\section{Problem Setting}

Operator learning aims to approximate the solution operator $\mathcal{G}$ that maps input functions to output fields of a PDE:
\begin{equation}
    \mathcal{G} : \mathcal{A}(\Omega) \rightarrow \mathcal{U}(\Omega), \quad a \mapsto u.
\end{equation}
In this paper, we consider elliptic boundary value problems (BVPs) of the general form
\begin{equation}
    \mathcal{L}(a(\mathbf{x}), u(\mathbf{x})) = f(\mathbf{x}) \quad \text{in } \Omega, \qquad \mathcal{B}u = g \quad \text{on } \partial\Omega,
\end{equation}
where $\mathcal{L} : \mathcal{A} \times \mathcal{U} \rightarrow \mathcal{F}$ is an elliptic differential operator and $\mathcal{B}$ a boundary operator on $\partial\Omega$. The functions $a \in \mathcal{A}(\Omega)$, $u \in \mathcal{U}(\Omega)$, and $f \in \mathcal{F}(\Omega)$ belong to appropriate function spaces over $\Omega \subset \mathbb{R}^d$, with $a$ a spatially varying coefficient or forcing term, $u$ the corresponding solution field, $f$ a source field, and $g$ the boundary data.

In practice, the fields can only be observed at finitely many coordinates. Denoting the discretely sampled input and output fields of the $i$-th training pair as $\hat{a}^{(i)}$ and $\hat{u}^{(i)}$, a family of neural networks $\{\mathcal{G}_\theta\}_{\theta \in \Theta}$ parameterized by model-parameters $\theta$ is trained by minimizing the loss function
\begin{equation}
    \tilde{\theta} = \operatorname*{arg\,min}_{\theta \in \Theta} \frac{1}{N} \sum_{i=1}^{N} \mathfrak{L}\left(\mathcal{G}_\theta(\hat{a}^{(i)}),\, \hat{u}^{(i)}\right),
\end{equation}
where $\mathfrak{L}$ is typically the mean squared error.

The learned forward-map $\mathcal{G}_{\tilde\theta}$ is generally not equivalent to the true operator $\mathcal{G}$, but is only an approximation over a subset $\tilde{\mathcal{A}} \subsetneq \mathcal{A}$ and $\tilde{\mathcal{U}} \subsetneq \mathcal{U}$.
These effective domains are shaped by the training distribution and by the properties of the forward map itself.
Generalization beyond $\tilde{\mathcal{A}}$, i.e., to input functions that differ in character, or statistical properties from those seen during training, cannot be guaranteed. Even when the inputs are drawn from a well-explored region of $\mathcal{A}$ the operator $\mathcal{G}$ may map non-uniformly onto $\mathcal{U}$, inducing a heavy-tailed output distribution in which rare solution structures are underrepresented.

\section{Methodology}
\subsection{Data-Generating Process}\label{sec:datagen_proc}
In the following, we propose a data-generating process (denoted as $\mathcal{P}$) to construct smooth signals with controllable properties. We want to be able to control structure sizes, data ranges and also be able generates non-stationary signals, exhibiting spatially varying correlation length scales. We define all fields on the flat unit torus $\mathbb{T}^2$, discretized on an $n \times n$ grid. Let $\mathcal{N}$ denote i.i.d. $\mathcal{N}(0, 1)$ noise on the grid, and let $\mathcal{K}_{\sigma}$ be the cyclic convolution with a Gaussian kernel of width $\sigma$. We define the following steps:
\paragraph{Stage 1 - Scale field $\sigma_s(\mathbf{x})$:}

If a stationary signal is needed, then this step can be omitted and $ \sigma_{s} $ can be just set to a constant
$ \sigma_{s}(\mathbf{x}) = \sigma_{s, 0} $.

For a non-stationary signal, we first sample $\zeta_s \sim \mathcal{N}$ and smooth it via $\tilde{\sigma}_s(\mathbf{x}) = \mathcal{K}_{\sigma_{\mathrm{ns}}}(\zeta_s)$, where $\sigma_{\mathrm{ns}}$ is a characteristic length scale of the non-stationarity. After this we min-max scale it onto $[\sigma_{\mathrm{s, min}}, \sigma_{\mathrm{s, max}}]$:
\begin{equation}
    \sigma_s(\mathbf{x}) = \sigma_{\mathrm{s, min}} + (\sigma_{\mathrm{s, max}} - \sigma_{\mathrm{s, min}}) \frac{\tilde{\sigma}_s(\mathbf{x}) - \mathrm{min} \, \tilde{\sigma}(\mathbf{x})}{\mathrm{max} \, \tilde{\sigma}(\mathbf{x}) - \mathrm{min} \, \tilde{\sigma}(\mathbf{x})}
\end{equation}

\paragraph{Stage 2 - Constructing the signal:}
Sample $\eta \sim \mathcal{N}$ and apply the Gaussian convolution
\begin{equation}
a(\mathbf{x}) = \int_{\mathbb{T}^2} k_{\sigma_s(\mathbf{x})} \left(\mathbf{x}-\mathbf{x}'\right) \eta(\mathbf{x}') \, \mathrm{d}\mathbf{x}'
\end{equation}
with a (possibly spatially varying) kernel $k_{\sigma_s(\mathbf{x})}(\cdot)$.
\paragraph{Stage 3 - Thresholding, jump and smoothing (optional):}
To introduce a smooth jump into the signal,  we first segment it into areas of two possible values by
\begin{equation}
    a(\mathbf{x}) \gets a_{\mathrm{min}} + (a_{\mathrm{max}} - a_{\mathrm{min}}) H\left(a(\textbf{x}) - \tau \right)
\end{equation}
where $H(\cdot)$ is the Heaviside function. After this, we smooth the resulting signal again by another convolution
\begin{equation}
    a(\mathbf{x}) \gets \mathcal{K}_{\sigma_{\mathrm{jump}}}(a(\mathbf{x}))
\end{equation}

\paragraph{Stage 4 - Amplitude modulation (optional):}
As another optional step we introduce a multiplicative amplitude modulation, by a signal of a big length scale $\sigma_{\mathrm{mod}} > \sigma_s$. With $\xi \sim \mathcal{N}$
\begin{equation}
    a(\mathbf{x}) \gets a(\mathbf{x}) \mathcal{K}_{\sigma_{\mathrm{mod}}}(\xi)
\end{equation}
\subsection{Equations}
Elliptic partial differential equations are a fundamental tool in science and engineering. They describe the steady-state of a wide range of phenomena, such as heat conduction, electrostatics, diffusion and linear elasticity. In this paper, we focus on the operator and not the geometry and set all problems on the unit square.
\subsubsection{Poisson-Type Equation}
\begin{equation}\label{eq:poissonseq}
    -\nabla \cdot \left(a(\mathbf{x})\, \nabla u(\mathbf{x})\right) = 1 
    \quad \text{in } \Omega = [0,1]^2, \qquad u = 0 \quad \text{on } \partial\Omega.
\end{equation}
with $a \in L^\infty(\Omega)$ and by the Lax-Milgram theorem a solution $u$ exists uniquely in the Sobolev space $u \in H^1_0(\Omega)$ and is H\"older-continuous by elliptic regularity. 
\subsubsection{Helmholtz Equation (Interior, Absorbing)}
\begin{equation}\label{eq:helmholtzeq}
    -\Delta u(\mathbf{x}) - (a(\mathbf{x}) + i a_i)^2\, u(\mathbf{x}) = f(\mathbf{x})
    \quad \text{in } \Omega = [0,1]^2, \qquad u = 0 \quad \text{on } \partial\Omega,
\end{equation}
where $a \in L^\infty(\Omega)$ is a spatially varying real wavenumber field, 
$a_i \geq 0$ is a constant absorption coefficient, and $f$ is a centered Gaussian 
excitation with a width $ \sigma $. The solution $u \in H^1_0(\Omega; \mathbb{C})$ is complex-valued.
The quantity of interest is the map $a \mapsto |u|$.

\subsection{Evaluation Tools}

\subsubsection{Empirical local Lipschitz analysis} 
To quantify how sensitively a forward operator $\mathcal{G}$ amplifies input perturbations, we introduce a procedure we call empirical local Lipschitz analysis. The key quantity of interest is a local amplification factor: for some base input function $a_0 \in \mathcal{A}$ and a perturbation $\delta \in \mathcal{A}$, we study the ratio of relative output deviation to relative input deviation as a function of perturbation magnitude.

Let $\mathcal{P}$ denote a data-generating process over input functions as defined in Section~\ref{sec:datagen_proc}.
For a single trial, we proceed as follows:
\begin{enumerate}
    \item Sample a base input $a_0 \sim \mathcal{P}$ and another independent realization $\tilde{a} \sim \mathcal{P}$. Define the centered perturbation as
          \begin{equation}\label{eq:perturbation_direction}
              \delta \;=\; \tilde{a} - \operatorname{mean}(\tilde{a}),
          \end{equation}
          where $\operatorname{mean}(\tilde{a})$ denotes the spatial mean over $\mathcal{D}$
    \item For a range of scaling factors
          $\varepsilon \in \{\varepsilon_1, \ldots, \varepsilon_K\}$ with
          $0 < \varepsilon_1 < \cdots < \varepsilon_K \ll 1$, calculate the
          perturbed input functions
          \begin{equation}\label{eq:perturbed_input}
              a_\varepsilon \;=\; a_0 + \varepsilon\, \delta.
          \end{equation}
    \item Compute the corresponding solutions $u_0 = \mathcal{G}(a_0)$ and $u_\varepsilon = \mathcal{G}(a_\varepsilon)$ via some classical solver (e.g., FEM), and evaluate the relative input and output deviations
          \begin{equation}\label{eq:relative_deviations}
              r_{\mathrm{in}}(\varepsilon)
                  \;=\; \frac{\|a_\varepsilon - a_0\|_{L^2(\mathcal{D})}}{\|a_0\|_{L^2(\mathcal{D})}}
                  \;=\; \varepsilon\,\frac{\|\delta\|_{L^2(\mathcal{D})}}{\|a_0\|_{L^2(\mathcal{D})}},
              \qquad
              r_{\mathrm{out}}(\varepsilon)
                  \;=\; \frac{\|u_\varepsilon - u_0\|_{L^2(\mathcal{D})}}{\|u_0\|_{L^2(\mathcal{D})}}.
          \end{equation}
    \item In the regime where $r_{\mathrm{out}}$ depends approximately linearly
          on $r_{\mathrm{in}}$, we estimate the slope
          \begin{equation}\label{eq:amplification_factor}
              \lambda(a_0, \delta)
                  \;=\; \frac{\mathrm{d}\, r_{\mathrm{out}}}{\mathrm{d}\, r_{\mathrm{in}}}
                  \;\approx\;
                  \frac{r_{\mathrm{out}}(\varepsilon)}{r_{\mathrm{in}}(\varepsilon)}
          \end{equation}
          via linear regression over the selected range of $\varepsilon$.
\end{enumerate}

We call $\lambda(a_0, \delta)$ the \emph{empirical local amplification factor} at $a_0$ in direction $\delta$. It can be interpreted as an empirical, direction-dependent local Lipschitz constant of $\mathcal{G}$ measured in relative norms: a value $\lambda > 1$ indicates that the operator amplifies perturbations, while $\lambda < 1$ indicates contraction.

To obtain a statistical characterisation of operator sensitivity for a given dataset, we repeat the above procedure for $M$ independent trials $(a_0^{(m)}, \delta^{(m)})_{m=1}^{M}$, each drawn independently from $\mathcal{P}$, and report aggregate statistics (mean, variance) of $\{\lambda^{(m)}\}_{m=1}^{M}$.

\subsubsection{Spectral centroid}

To characterize the frequency content of a signal $a \colon \mathcal{D} \to \mathbb{R}$ discretized on an $N \times N$ grid, we use the spectral centroid, a standard measure from signal processing that captures the magnitude-weighted mean frequency of a signal. In two spatial dimensions, it is defined as
\begin{equation}\label{eq:spectral_centroid}
\bar{k}(a) =
\sum_{\mathbf{k}} \|\mathbf{k}\|\, |\hat{a}(\mathbf{k})|
\left(\sum_{\mathbf{k}} |\hat{a}(\mathbf{k})|\right)^{-1}
\end{equation}
where $\hat{a}(\mathbf{k})$ denotes the 2D discrete Fourier transform of $a$. The wave vector is defined as $\mathbf{k} = (k_x, k_y)\,,$
Higher values of $\bar{k}$ indicate a shift of spectral mass towards higher frequencies.

\section{The ELADO Benchmark}
All datasets were generated by finite element simulations using \texttt{FEniCSx}/\texttt{dolfinx}.
For the Poisson-Type equation (Equation \ref{eq:poissonseq}), the unit-square domains were discretized by a structured triangular mesh of $511 \times 511$ square cells (split into two triangles). The per-axis resolution of the sampled coefficient fields is therefore 512. Both the solutions $u$ and the coefficients $a$ were represented in the space of continuous piecewise-linear Lagrange elements ($P_1$), so that $a$ is interpolated from its nodal values. The resulting linear systems were solved with the \texttt{PETSc} library using the Conjugate Gradient method  with a HYPRE algebraic multigrid preconditioner.
For the Helmholtz equation (Equation \ref{eq:helmholtzeq}), the unit-square domains were discretized by a structured triangular mesh of $255 \times 255$ square cells (split into two triangles). The per-axis resolution of the sampled coefficient fields is $256$. The complex-valued problem was transformed into an equivalent coupled real system for the real and the imaginary parts $(u_r, u_i)$. The wavenumber $k$ was again interpolated from the sampled field into a continuous piecewise-linear ($P_1$) Lagrange space, while the coupled solution $(u_r, u_i)$ was discretized with $P_2$ elements using a Lagrange basis with nodes at the three vertices and the edge midpoints of each triangle. The resulting system was solved by a direct LU factorization  (\texttt{MUMPS}) through \texttt{PETSc}.
\subsection{Heavy-Tailed Solution Distributions Arising from Light-Tailed Coefficient Field Distributions}\label{seq:heavy_tailed1}

\begin{figure}[t]
\centering
\includegraphics[width=0.8\textwidth]{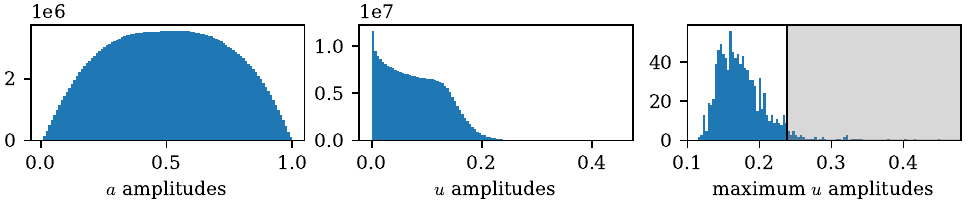}
\caption{Histograms for \dset{A1-train}. The values of the input field $a$ are short-tailed, while the values of the output field $u$ are heavy-tailed. The right panel shows the distribution of per-sample maxima of $u$; the shaded region marks samples exceeding the $95$th percentile. The \dset{A1-tail} dataset is constructed by drawing exclusively from this region.} \label{fig:51_sparse_oversampling}
\end{figure}

The datasets defined in this section demonstrate that even if the input field produced by the data-generating process exhibits light-tailed amplitude behavior, the resulting solution field can develop a heavy-tailed amplitude distribution. Such situations are particularly challenging for learning algorithms, as the tail of the solution distribution is severely undersampled. Moreover, there is in general no straightforward way to determine  which samples will give rise to extreme solution values from the input field alone.
The data-generating process $\mathcal{P}_1$ (as described in Section \ref{sec:datagen_proc}) used in this section is stationary and does not employ amplitude modulation. We set $\sigma_{s,0} = 0.04$, $\sigma_{\mathrm{jump}} = 0.06$, $a_{\mathrm{min}} = 0.01$, $a_{\mathrm{max}} = 1.0$, and $\tau = 0$. The PDE-problem is the forward-map of Poisson's equation as described in Equation~\ref{eq:poissonseq}.
We generate training (\dset{A1-train}) and test (\dset{A1-test}) datasets of 1000 samples each. In addition, we construct a \emph{tail dataset} (\dset{A1-tail}) of 500 samples whose maximum solution value exceeds the $95$th percentile of the training set. This ensures that sufficient samples are available to evaluate model performance in the sparse tail of the solution distribution (see Figure~\ref{fig:51_sparse_oversampling}). 
% As we will see, however, the maximum amplitude of the solution is only a moderate predictor of the relative $L^2$ prediction error.

\begin{figure}[t]
\centering
\includegraphics[width=0.7\textwidth]{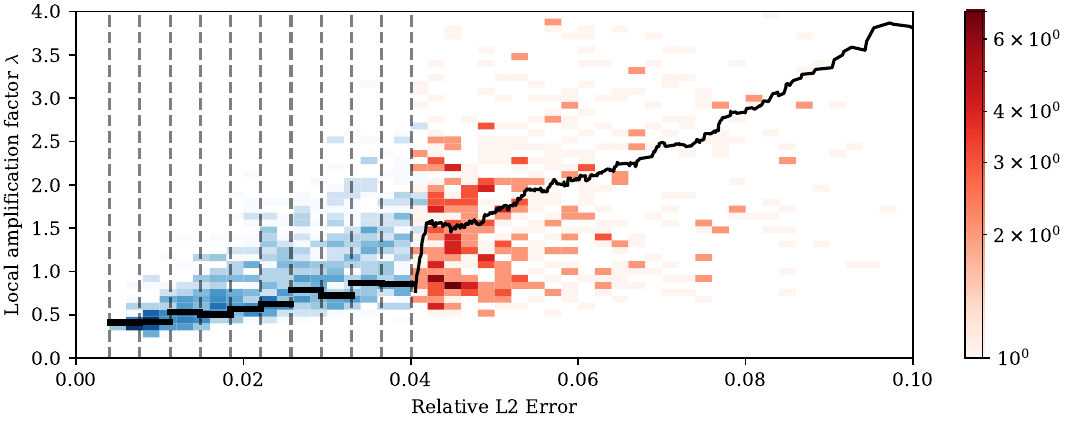}
\caption{Joint distribution of the relative $L^2$ prediction error of a trained FNO model and the local amplification factor $\lambda$ for samples from \dset{A1-train}, shown as a 2D histogram (log-scale color). The black curve shows the conditional mean of $\lambda$: within the oversampling range (blue) it is averaged over the fixed bins; beyond it, a sliding-window average is used. Dashed vertical lines indicate the bin boundaries used to construct the oversampled datasets (\dset{A1-os-train}, \dset{A1-os-test}), with each bin containing an equal number of samples. The bulk of the distribution is concentrated at low error and low $\lambda$ (blue region); samples in the tail of the error distribution exhibit increased amplification factors, indicating that prediction difficulty correlates with some intrinsic sensitivity-property of the forward operator itself.} \label{fig:51_os_strategy}
\end{figure}

We further define a dataset to investigate how effectively oversampling is at compensating for the sparsity in the solution distribution. To this end, we adopt a model-in-the-loop strategy. We take the relative $L^2$ prediction errors of a trained FNO model (trained with \dset{A1-train}) and partition the interval between the smallest error and its $99$th percentile into 10 equally spaced bins. We then repeatedly simulate new samples, predict with the FNO model, and sort each sample into the corresponding bin until every bin contains 100 samples. Both a training (\dset{A1-os-train}) and a test (\dset{A1-os-test}) set are constructed in this way (see Figure~\ref{fig:51_os_strategy}). To quantify the degree to which this oversampling strategy mitigates the effects of distributional sparsity, we compare the relative $L^2$ prediction error on the following setups: We perform a training on the "base" and the "os" training datasets and also evaluate on the two test datasets. We call these four cases ($\textsf{train}\!\to\!\textsf{test}$): $\textsf{base}\!\to\!\textsf{base}$, $\textsf{base}\!\to\!\textsf{os}$, $\textsf{os}\!\to\!\textsf{base}$, and $\textsf{os}\!\to\!\textsf{os}$. Let $E(\cdot)$ be the mean relative $L^2$ error of those cases on the oversampling interval, then we define three diagnostic ratios to characterize the interaction between distributional sparsity and learnability:
\begin{itemize}
    \item $E(\textsf{base}\!\to\!\textsf{os}) / E(\textsf{base}\!\to\!\textsf{base}) = 1.77$ quantifies \textbf{tail sensitivity}: the factor by which prediction error increases when a model trained on the natural (heavy-tailed) distribution is evaluated on a uniformly sampled test dataset.
    \item $1 - (E(\textsf{os}\!\to\!\textsf{os}) / E(\textsf{base}\!\to\!\textsf{os})) = 0.30$ quantifies \textbf{tail learnability}: the factor of the tail error that is recoverable by oversampling. A value of $0$ means oversampled training recovered none of the tail error, a value of $1$ would mean all tail error was eliminated. A negative value means oversampled training actively hurts the model.
    \item $E(\textsf{os}\!\to\!\textsf{base}) / E(\textsf{base}\!\to\!\textsf{base}) = 1.23$ quantifies \textbf{oversampling cost}: the degradation in overall accuracy induced by redistributing training towards tail samples. This indicates, that improving tail performance comes at the expense of overall accuracy.
\end{itemize}
Taken together, these ratios show that oversampling may provide a partial but not complete remedy: tail errors decrease, yet bulk accuracy suffers.
The tail learnability is significantly smaller than $1$ and the elevated local amplification factors in the tail region (Figure \ref{fig:51_os_strategy}) point to the forward map itself as a source of these hard samples, suggesting reweighting the training distribution alone would be insufficient.
We note that the oversampled datasets (\dset{A1-os-train}, \dset{A1-os-test}) were constructed using a single FNO model, which introduces an architecture-dependent bias into these datasets. Due to the computational expense of this procedure, we did not perform it for the other architectures. However, the oversampled datasets remain informative for other models when the per-sample prediction errors (\dset{A1-train}, \dset{A1-test}) correlate across architectures (verified via correlation coefficient).

\begin{figure}[t]
\centering
\includegraphics[width=0.7\textwidth]{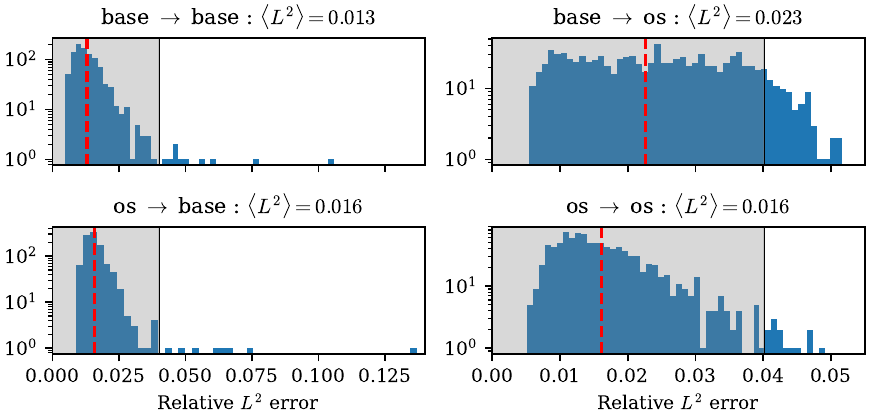}
\caption{Histograms of the relative $L^2$ prediction error (log-scaled y-axis) for the four train $\to$ test configurations of the oversampling experiment. The red dashed lines marks the mean relative error $\langle L^2 \rangle$ (reported in each tile); the shaded regions show the oversampling interval. The errors were only evaluated on this interval.} \label{fig1}
\end{figure}

\subsection{Spectral Distribution Shift of the Input Data}
The datasets defined in this section are designed to probe how prediction accuracy degrades when the input field is shifted out of distribution relative to the training data. The data-generating process $\mathcal{P}_2$ is identical to that in Section~\ref{seq:heavy_tailed1}, except for the values of $\sigma_{s,0}$ and $\sigma_{\mathrm{jump}}$.
For the first dataset we set $\sigma_{s,0} = 0.02$ and $\sigma_{\mathrm{jump}} = 0.0267$ and generate 1000 samples (\dset{B1-test}). This shifts the spectral content of the input field toward higher frequencies compared to the training distribution. By retaining the same $a_{\mathrm{min}}$ and $a_{\mathrm{max}}$ as in the previous section, we ensure that the coefficient fields span the same value range. This isolates the effect of spectral mismatch: any change in test accuracy can be attributed to the frequency shift rather than to a change in amplitude scale. The intended use is to evaluate the relative $L^2$ prediction error on this dataset using models trained on the training set from Section~\ref{seq:heavy_tailed1}.
For the second dataset (\dset{B1-ramp}) we vary $\sigma_{s,0}$ and $\sigma_{\mathrm{jump}}$ continuously over the intervals
\begin{equation}
    \tilde{\sigma}_{s,0} \in [0.01,\, 0.07], \qquad \tilde{\sigma}_{\mathrm{jump}} \in [0.01,\, 0.11],
\end{equation}
sampling 2000 equidistant points along this range. This dataset serves as an analytical tool for visualizing how prediction accuracy evolves as the spectral content of the input field is gradually shifted away from the training distribution.

\subsection{Heavy-Tailed Distributions in the Frequency Domain of Solutions, Arising From Light-Tailed Coefficient Field Distributions}

\begin{figure}[t]
\centering
\includegraphics[width=0.8\textwidth]{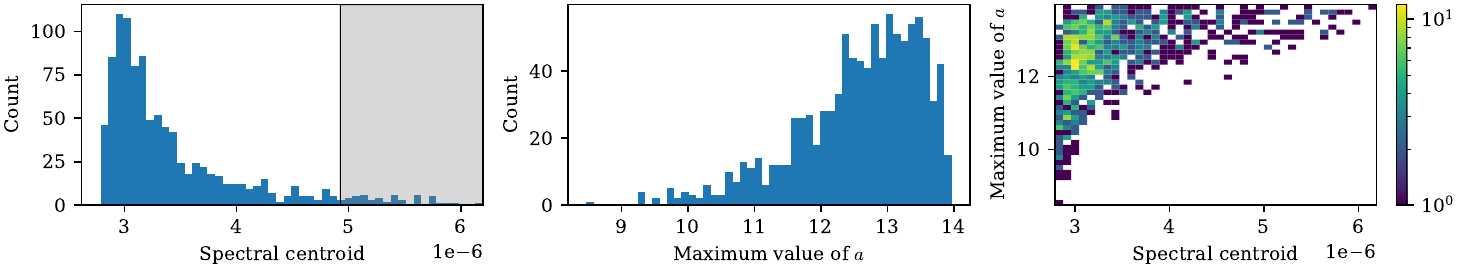}
\caption{Distributional characteristics of \dset{C1-train}. Left: distribution of spectral centroid values; the shaded region marks samples exceeding the $95$th percentile, from which \dset{C1-tail} is constructed. Center: distribution of per-sample maximum values of $u$. Right: joint distribution of spectral centroid values and maximum value of $a$ (log-scale color). The spectral centroid distribution is right-skewed, producing a heavy tail in the frequency domain. The joint distribution reveals no strong correlation between maximum input values and spectral content, indicating that high-frequency outliers cannot be identified from the amplitude of $a$ alone.} \label{fig:53_dists}
\end{figure}

The datasets defined in this section mirror the structure of Section~\ref{seq:heavy_tailed1}. Here, however, the heavy-tailed behavior manifests in the frequency domain of the solution rather than in its amplitude. We employ the Helmholtz equation as described in Equation~\ref{eq:helmholtzeq} together with a stationary, unmodulated data-generating process $\mathcal{P}_3$. The remaining parameters are $a_{\mathrm{min}} = 6$, $a_{\mathrm{max}} = 14$, $\tau = 3.84$, $\sigma_{s,0} = 0.0684$, $\sigma_{\mathrm{jump}} = 0.0684$, and $a_i = 1$. These settings are chosen so that the bulk of the solutions contains approximately half a spatial period, with occasional outliers exhibiting up to roughly one full period. We quantify this frequency content using the spectral centroid defined in Equation~\ref{eq:spectral_centroid}.
We generate training (\dset{C1-train}) and test (\dset{C1-test}) datasets of 1000 samples each. Additionally, we construct a tail dataset (\dset{C1-tail}) by oversampling 500 samples whose spectral centroid exceeds the $95$th percentile of the training set (see Figure~\ref{fig:53_dists}).

\subsection{Spectral Distribution Shift of the Output Data}
The datasets which we define in this section mirror those in the second subsection. We want to adjust the input signal (without changing its value range)
to change the frequency response. We achieve this by varying $\tau$ and create two datasets. The first (\dset{D1-test}) has 500 samples and has $\tau=0$, and the second (\dset{D1-ramp}) one is a ramp of 2000 equidistant data points on the range
\begin{equation}
    \tilde{\tau} \in \left[-2.56, 5.12 \right].
\end{equation}

\subsection{High Local Amplification Factor (Sensitivity)}
Below, we define a datasets that is harder to learn than those in the previous sections. For this we generate 1000 training and 1000 test samples each and call them (\dset{E1-train} and \dset{E1-test}). We again use the Helmholtz equation as defined in Equation~\ref{eq:helmholtzeq} with $a_i = 1$. For the data-generating process $\mathcal{P}_{5,a}$ we set $a_{\mathrm{min}} = 16$, $a_{\mathrm{max}} = 30$, $\tau = 0.03$, $\sigma_{s, 0} = 0.0684$, and $\sigma_{\mathrm{jump}} = 0.0684$. By doing so, the input signal has larger amplitudes than the signals in the previous sections, and therefore solutions with higher frequencies - giving the signal more complexity and therefore making it harder to learn. We perform a local Lipschitz analysis with 1000 simulations and find that the local amplification factor is in the range $4.2\pm0.6$. This value is significant larger than in the previous experiments and means that the output is highly sensitive to the input for the entire dataset and therefore poses a greater difficulty for operator learning models.

\subsection{Signal Complexity Effects}
In this subsection we define four datasets (train and test, 1000 samples each) (\dset{F1-train}, \dset{F1-test}, ..., \dset{F4-test}) with increasing input-signal complexity. For this we define four data-generating processes:

\begin{itemize}
    \item $\mathcal{P}_{7a}$ coarse scaled: No jump, no modulation, stationary: $\sigma_0 = 0.08$
    \item $\mathcal{P}_{7b}$ fine scaled: No jump, no modulation, stationary: $\sigma_0 = 0.03$
    \item $\mathcal{P}_{7c}$ multi-scaled: No jump, no modulation, non-stationary $\sigma_{\mathrm{s, min}} = 0.03$, $\sigma_{\mathrm{s, max}} = 0.08$, $\sigma_{\mathrm{ns}} = 0.15$
    \item $\mathcal{P}_{7d}$ No jump, modulated, non-stationary: $\sigma_{\mathrm{s, min}} = 0.03$, $\sigma_{\mathrm{s, max}} = 0.08$, $\sigma_{\mathrm{ns}} = 0.15$, and $\sigma_{\mathrm{mod}} = 0.2$
\end{itemize}
After generation, we then standardize the signals per dataset via
\begin{equation}
    a(\mathbf{x}) \gets \alpha \, \frac{a(\mathbf{x}) - \mu_a}{\sigma_a}
\end{equation}
with a factor we chose as $\alpha=1.5$. As final step we apply $a(\mathbf{x}) \gets e^{a(\mathbf{x})}$, which ensures positivity of the coefficient field and introduces high contrast ratios between local maxima and minima of $a$.

With these datasets the following observations can be made (not absolute, but compared relatively to each other):
\begin{itemize}
    \item $\mathcal{P}_{7a}$ \& $\mathcal{P}_{7b}$: Does increased scale complexity worsen model accuracy?
    \item $\mathcal{P}_{7b}$ \& $\mathcal{P}_{7c}$: With a minimal signal scale deliberately chosen to be the same as in b: Does non-stationarity of the input signal worsen model accuracy?
    \item $\mathcal{P}_{7b,c}$ \& $\mathcal{P}_{7d}$: Does a amplitude modulation worsen model accuracy?% (more variance in amplitudes)? 
\end{itemize}

\section{Results}
We evaluated four common operator learning architectures on our benchmark. We use their respective default settings and keep them across all experiments. In all cases we trained and evaluated on regular grids of $128 \times 128$ grid points (sub-sampling the Poisson's equation datasets by a factor of $4$ and the Helmholtz equation datasets by a factor of $2$). The architectures used are DeepONet ($4.5\,\mathrm{M}$ parameters), FNO ($139\,\mathrm{M}$ parameters), OFormer ($2.6\,\mathrm{M}$ parameters), and Galerkin-Transformer ($2.2\,\mathrm{M}$ parameters). Importantly, our objective is not to compare accuracies of these models directly or to find a best model, but to interpret the relative changes.

When error intervals are given, these show the mean and standard deviation over 5 independent realizations of that experiment.

\subsection{Heavy-Tailed Solution Distributions}

Table~\ref{tab:heavy_tail_amp} reports the mean relative $L^2$ prediction errors for all models trained on \dset{A1-train} and evaluated under three conditions: the standard test set \dset{A1-test}, the tail-oversampled set \dset{A1-tail}, and the distribution-shifted set \dset{B1-test}.
First, all models degrade on \dset{A1-tail}, confirming that the samples from the heavy tail of the distribution of $u$ produces larger errors. However, the degree of tail-fragility varies notably: the Galerkin Transformer exhibits the highest tail fragility with a roughly $2.9 \times$ increase in error, despite being competitive with FNO on the standard benchmark. FNO, in contrast, maintains the lowest absolute error across all three conditions, suggesting a more consistent performance across difficulty levels.

Second, and perhaps more striking, the model ranking is not preserved under distribution shift. On \dset{A1-test}, the Galerkin Transformer outperforms OFormer, but on \dset{B1-test} this ordering reverses, representing an approximately $4.7 \times$ degradation for the Galerkin Transformer. This instability in ranking illustrates exactly the limitation of reporting metrics on a single test distribution.

\begin{table}
\centering
\caption{Mean relative $L^2$ prediction errors of different models trained on \dset{A1-train} and evaluated on \dset{A1-test}, \dset{A1-tail}, and \dset{B1-test}}\label{tab:heavy_tail_amp}
\begin{tabular}{@{} l @{\hspace{1em}} S[table-format=1.2(2), table-space-text-post={\%}]
                      @{\hspace{1em}} S[table-format=1.2(2), table-space-text-post={\%}] 
                      @{\hspace{1em}} S[table-format=1.2(2), table-space-text-post={\%}] @{}}
\toprule
{Model}       & {\dset{A1-test}} & {\dset{A1-tail}} & {\dset{B1-test}} \\
\midrule
DeepONet      & 12.5(1) \%       & 19.8(1) \%       & 17.0(2) \%  \\
FNO           & 1.6(4) \%        & 3.6(7) \%        & 4.3(10) \%  \\
OFormer       & 2.49(03) \%      & 4.73(01) \%      & 5.9(02) \%  \\
Galerkin T.   & 1.98(03) \%      & 5.66(09) \%      & 8.9(03) \%  \\
\bottomrule
\end{tabular}
\end{table}

Table~\ref{tab:heavy_tail_oversamp} presents the full oversampling experiments. Since the oversampled dataset was constructing using FNO's per-sample errors, we first assess transferability with the calculation of the correlation of the per-sample errors (on the heavy-tailed interval). OFormer and Galerkin Transformer exhibit strong correlation with FNO, indicating that per-sample difficulty is shared across these architectures. DeepOnet, with a correlation of $0.46$, is excluded from further analysis. Among the remaining models, OFormer and Galerkin Transformer both exhibit a tail sensitivity of $1.3$, notably lower than FNO with $1.8$. This suggests that these architectures architectures are less affected by evaluation on the uniformly sampled test set. The differs across models: FNO recovers approximately $30 \%$ and Galerkin Transformer recovers approximately $20 \%$ of the tail error by oversampling. OFormer has a tail learnability of $0.0$, meaning that oversampling does not provide any improvement in the tail region. The oversampling cost factor for FNO and OFormer architectures is larger than $1$. However for the Galerkin Transformer architecture it is $0.7$, indicating that redistributing training towards tail samples actually improves its overall accuracy on the natural distribution.

\begin{table}%[t]
\centering
\caption{Oversampling properties of different models}\label{tab:heavy_tail_oversamp}
\begin{tabular}{@{} l @{\hspace{1em}}
                    | l @{\hspace{1em}} |
                    l @{\hspace{1em}}
                    l @{\hspace{1em}}
                    l @{\hspace{1em}} 
                    @{}}
\toprule
{Model}                                & FNO     & DeepONet & OFormer    & Galerkin-T  \\
\midrule
FNO-corr.                              & 1.0     & 0.46     & 0.81       & 0.84 \\
$ \mathrm{base} \to \mathrm{base} $    & 1.3 \%  & -        & 2.3 \%     & 2.3 \%  \\
$ \mathrm{base} \to \mathrm{os} $      & 2.3 \%  & -        & 2.9 \%     & 2.9 \%  \\
$ \mathrm{os} \to \mathrm{base} $      & 1.6 \%  & -        & 2.5 \%     & 1.7 \%  \\
$ \mathrm{os} \to \mathrm{os} $        & 1.6 \%  & -        & 2.9 \%     & 2.3 \%  \\
\hline
tail sensitivity                       & 1.8     & -        & 1.3        & 1.3     \\
tail learnability                      & 0.3     & -        & 0.0        & 0.2     \\
oversampling cost                      & 1.2     & -        & 1.1        & 0.7    \\
\bottomrule
\end{tabular}
\end{table}

Table~\ref{tab:heavy_tail_amp:prediction:error:1} reports the corresponding results for the heavy-tailed frequency setting. Compared to the amplitude case, errors are higher across all models and datasets, indicating that the this specific forward-map is harder to learn then the previous one. The fragility ratios are very similar among FNO, OFormer, and the Galerkin Transformer, all falling in the range $2.3-2.4 \times$.

This contrasts with the amplitude setting, where Galerkin Transformer exhibited notably higher fragility than the other models, and suggests that the difficulty of a heavy tail in the frequency-domain is more intrinsic to the problem than to the architecture.

Distribution shift is severe for all models, with degradation ratios in the range $3.8 - 4.9 \times$.

\begin{figure}%[t]
\centering
\includegraphics[width=0.55\textwidth]{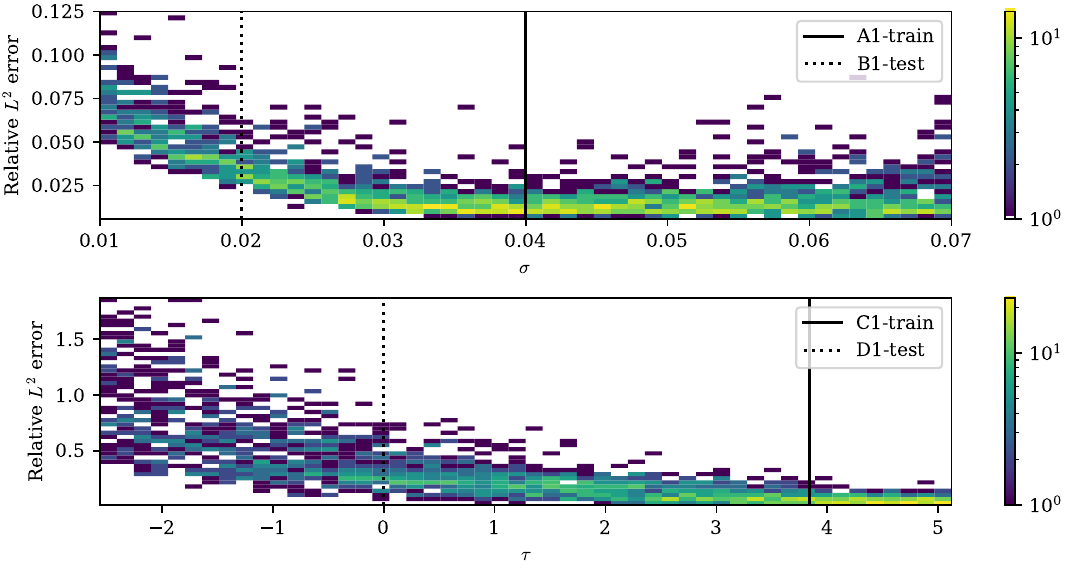}
\caption{Visualizations of datasets \dset{B1-ramp} and \dset{D1-ramp}. The dataset parameters $\sigma$ and $\tau$ were gradually shifted around their values used in \dset{A1-train} and \dset{B1-train}} \label{fig:abcd_ramps}
\end{figure}

Figure~\ref{fig:abcd_ramps} shows a visualization of datasets \dset{B1-ramp} and \dset{D1-ramp}. These were used as a diagnostic tool to find the parameters of datasets \dset{B1-test} and \dset{D1-test}.

\begin{table}
\centering
\caption{Relative $L^2$ prediction errors (mean $\pm$ std) on the
heavy-tailed frequency datasets.}\label{tab:heavy_tail_amp:prediction:error:1}
\begin{tabular}{@{} l @{\hspace{1em}} S[table-format=1.2(2), table-space-text-post={\%}]
                      @{\hspace{1em}} S[table-format=1.2(2), table-space-text-post={\%}] 
                      @{\hspace{1em}} S[table-format=1.2(2), table-space-text-post={\%}] @{}}
\toprule
{Model}       & {\dset{C1-test}} & {\dset{C1-tail}} & {\dset{D1-test}} \\
\midrule
DeepONet      & 24.0(06) \%      & 39.0(06) \%      & 90.0(30) \%  \\
FNO           & 7.6(03) \%      & 18.1(05) \%      & 35.0(20) \%  \\
OFormer       & 7.5(01) \%       & 17.2(01) \%       & 36.6(01) \%  \\
Galerkin T.   & 8.8(01) \%       & 20.0(01) \%       & 41.2(10) \%  \\
\bottomrule
\end{tabular}
\end{table}

\subsection{High Local Amplification Factor}
The results for the PDE setting with the high amplification factors (\dset{E1-test}) are DeepONet: $35.0 \pm 0.5 \%$, FNO: $19.1 \pm 0.3 \%$, OFormer: $23.8 \pm 0.1 \% $, and Galerkin T.: $22.0 \pm 0.1 \% $. The errors are elevated compared to the other benchmarks, supporting our claim that datasets with high amplification factors pose a challenge for operator learning architectures.

\subsection{Signal Complexity}

Table~\ref{tab:heavy_tail_amp:prediction:error:3} reports results across increasing levels of input signal complexity. Most notably, the Galerkin Transformer failed to converge on all four datasets, producing numerical instabilities (NaN) early in training. 

Among the remaining models, the ranking of FNO and OFormer is fully preserved across all four datasets. Errors increase monotonically at each stage: fine-scaled signals are harder to learn than coarse-grained ones; introducing multiple scales (non-stationarity) increases the difficulty further, even when the finest scale is no finer than the single-scale case; and additional amplitude modulation also increases prediction error. DeepONet does not fully follow the same ranking, though this is probably attributable to its already elevated baseline error.

\begin{table}%[t]
\centering
\caption{Mean relative $L^2$ prediction errors for the datasets with increasing complexity.}\label{tab:heavy_tail_amp:prediction:error:3}
\begin{tabular}{@{} l @{\hspace{1em}} S[table-format=1.2(2), table-space-text-post={\%}]
                      @{\hspace{1em}} S[table-format=1.2(2), table-space-text-post={\%}] 
                      @{\hspace{1em}} S[table-format=1.2(2), table-space-text-post={\%}] 
                      @{\hspace{1em}} S[table-format=1.2(2), table-space-text-post={\%}] @{}}
\toprule
{Model}       & {\dset{F1-test}} & {\dset{F2-test}} & {\dset{F3-test}} & {\dset{F4-test}} \\
\midrule
DeepONet      & 24.1(02) \%            & 21.7(01) \%      & 26.3(01) \%  & 31.6(03) \% \\
FNO           & 5.4(03) \%             & 7.7(01) \%      & 8.5(02) \%  & 9.2(01) \%\\
OFormer       & 10.2(01) \%            & 12.4(01) \%      & 14.3(01) \%  & 16.2(01) \%\\
\bottomrule
\end{tabular}
\end{table}
\section{Conclusions}

In this paper, we introduced ELADO, a benchmark suite to expose and quantify failure modes of neural operator architectures for elliptic PDEs. ELADO focuses on controlled scenarios that may naturally arise in elliptic problems but remain underexplored in current benchmarks. Through a controllable data-generating process and datasets built around Poisson's equation and the Helmholtz equation, the benchmark isolates several distinct sources of difficulty relevant to operator learning.

Our experiments demonstrate several distinct issues:
\begin{compactenum}%[(1)]
\item Heavy-tailed behavior in the solution space can arise when the coefficient fields follow light tailed distributions, leading to rare but significant samples.
\item We further show that oversampling may only partially solve these issues and can even introduce new biases.
\item We also include datasets that introduce a controlled shift in the distribution of the input fields relative to the training distribution. The purpose of these datasets is not to demonstrate the existence of this effect, but to provide a controlled benchmark setting.
\item Another aspect explored in ELADO is operator sensitivity. We construct a dataset in which small perturbations of the input fields lead to comparatively large changes in the corresponding solutions, resulting in high empirical local amplification factors. On this dataset all evaluated neural operator architectures show substantially reduced prediction accuracy. These cases highlight regimes in which the underlying PDE operator is highly sensitive to input perturbations and therefore particularly challenging for current operator learning models.
\item Finally, we investigate how the structural complexity of the input fields influences prediction accuracy. By increasing the complexity of the coefficient fields (through changes in scale-size, non-stationarity, and through amplitude modulation) while keeping their overall amplitude range normalized, we observe a consistent but moderate degradation in model accuracy. This experiment isolates input signal complexity as an independent factor influencing performance and suggests that even controlled increases in structural complexity can affect the learnability of the corresponding solution operator.
\end{compactenum}

Taken together, these observations show that current evaluation practices based primarily on mean error metrics and relatively homogeneous datasets may obscure important failure modes of neural operator models.


\begin{thebibliography}{10}
\providecommand{\url}[1]{\texttt{#1}}
\providecommand{\urlprefix}{URL }
\providecommand{\doi}[1]{https://doi.org/#1}

\bibitem{azizzadenesheli2024neural}
Azizzadenesheli, K., Kovachki, N., Li, Z., Liu-Schiaffini, M., Kossaifi, J.,
  Anandkumar, A.: Neural operators for accelerating scientific simulations and
  design. Nature Reviews Physics  \textbf{6}(5),  320--328 (2024)

\bibitem{chen1995universal}
Chen, T., Chen, H.: Universal approximation to nonlinear operators by neural
  networks with arbitrary activation functions and its application to dynamical
  systems. IEEE transactions on neural networks  \textbf{6}(4),  911--917
  (1995)

\bibitem{gupta2022towards}
Gupta, J.K., Brandstetter, J.: Towards multi-spatiotemporal-scale generalized
  pde modeling. arXiv preprint arXiv:2209.15616  (2022)

\bibitem{Kovachki2021NeuralOL}
Kovachki, N.B., Li, Z.Y., Liu, B., Azizzadenesheli, K., Bhattacharya, K.,
  Stuart, A.M., Anandkumar, A.: Neural operator: Learning maps between function
  spaces. arXiv  \textbf{abs/2108.08481} (2021)

\bibitem{li2023fourier}
Li, Z., Huang, D.Z., Liu, B., Anandkumar, A.: Fourier neural operator with
  learned deformations for pdes on general geometries. JMLR  \textbf{24}(388),
  1--26 (2023)

\bibitem{li2020fourier}
Li, Z., Kovachki, N., Azizzadenesheli, K., Liu, B., Bhattacharya, K., Stuart,
  A., Anandkumar, A.: Fourier neural operator for parametric partial
  differential equations. arXiv preprint arXiv:2010.08895  (2020)

\bibitem{li2020multipole}
Li, Z., Kovachki, N., Azizzadenesheli, K., Liu, B., Stuart, A., Bhattacharya,
  K., Anandkumar, A.: Multipole graph neural operator for parametric partial
  differential equations. Proc. NeurIPS  \textbf{33},  6755--6766 (2020)

\bibitem{liu2021multiscale}
Liu, L., Cai, W.: Multiscale deeponet for nonlinear operators in oscillatory
  function spaces for building seismic wave responses. arXiv preprint
  arXiv:2111.04860  (2021)

\bibitem{liu2022causality}
Liu, L., Nath, K., Cai, W.: A causality-deeponet for causal responses of linear
  dynamical systems. arXiv preprint arXiv:2209.08397  (2022)

\bibitem{lu2019deeponet}
Lu, L., Jin, P., Karniadakis, G.E.: Deeponet: Learning nonlinear operators for
  identifying differential equations based on the universal approximation
  theorem of operators. arXiv preprint arXiv:1910.03193  (2019)

\bibitem{takamoto2024pdebenchextensivebenchmarkscientific}
Takamoto, M., Praditia, T., Leiteritz, R., MacKinlay, D., Alesiani, F.,
  Pflüger, D., Niepert, M.: Pdebench: An extensive benchmark for scientific
  machine learning (2024), \url{https://arxiv.org/abs/2210.07182}

\end{thebibliography}
\end{document}